\documentclass[12pt]{article}
\usepackage{amsfonts}
\usepackage{a4wide}
\usepackage{amsmath,amscd,amsthm,a4,amssymb}
\usepackage[hyperfootnotes=false]{hyperref}

\begin{document}

\begin{center}
{\Large \textbf{On shallow feedforward neural networks with inputs from a
topological space}}

\

\textsc{Vugar E. Ismailov} \

\bigskip

Institute of Mathematics and Mechanics, Baku, Azerbaijan

Center for Mathematics and its Applications, Khazar University, Baku,
Azerbaijan

{e-mail:} {vugaris@mail.ru}
\end{center}

\begingroup
\renewcommand{\thefootnote}{}
\footnotetext{\raggedright Published in \textit{Annals of Mathematics and Artificial Intelligence}. DOI: \href{https://doi.org/10.1007/s10472-026-10003-7}{10.1007/s10472-026-10003-7}.}
\endgroup

\smallskip

\textbf{Abstract.} We study feedforward neural networks with inputs from a
topological space (TFNNs). We prove a universal approximation theorem for
shallow TFNNs, which demonstrates their capacity to approximate any
continuous function defined on this topological space. As an application, we
obtain an approximative version of Kolmogorov's superposition theorem for
compact metric spaces.

\bigskip

\textit{Mathematics Subject Classifications:} 41A30, 41A65, 68T05

\smallskip

\textit{Keywords:} feedforward neural network, universal approximation
theorem, density, topological space, locally convex space, Tauber-Wiener function,
Kolmogorov's superposition theorem

\

\begin{center}
{\large \textbf{1. Introduction}}
\end{center}

Neural networks are fundamental to contemporary machine learning and
artificial intelligence, providing robust methods for tackling intricate
challenges. Among the different neural network designs, the \textit{%
multilayer feedforward perceptron} (MLP) is particularly prominent and
essential. The MLP is valued for its capability to model complex, nonlinear
functions and execute various tasks, including classification, regression,
and pattern recognition.

This architecture consists of a limited number of sequential layers: an
input layer at the beginning, an output layer at the end, and several hidden
layers in between. Information progresses from the input layer through the
hidden layers to the output layer. In this framework, each neuron in a layer
receives inputs from the previous layer, applies specific weights, adds a
bias, and then processes the result through an activation function. This
activation function introduces non-linearity, allowing the model to learn
and capture intricate patterns. The output from one layer's neurons serves
as the input for the neurons in the next layer, continuing this sequence
until the final output is generated by the output layer.

The most basic form of an MLP features just one hidden layer. In this setup,
each output neuron calculates a function expressed as
\begin{equation*}
\sum_{i=1}^{r}c_{i}\sigma (\mathbf{w}^{i}\cdot \mathbf{x}-\theta _{i}),\eqno%
(1.1)
\end{equation*}%
where $\mathbf{x}=(x_{1},...,x_{d})$ represents the input vector, $r$ is the
number of neurons in the hidden layer, $\mathbf{w}^{i}$ are \textit{weight
vectors} in $\mathbb{R}^{d}$, $\theta _{i}$ are \textit{thresholds}, $c_{i}$
are coefficients, and $\sigma $ is the \textit{activation function}, a real
univariate function.

The theoretical underpinning of neural networks is rooted in the \textit{%
universal approximation property} (UAP), sometimes referred to as the
density property. This principle states that a neural network with a single
hidden layer can approximate any continuous function over a compact domain
to any desired level of precision. Specifically, the set of all functions
defined in the format of equation (1.1), is dense in $C(K)$ for every
compact set $K\subset \mathbb{R}^{d}$. Here $C(K)$ represents the space of
real-valued continuous functions on $K$. This important result in neural
network theory is known as the \textit{universal approximation theorem}
(UAT).

Note that the inner product $\mathbf{w}^{i}\cdot \mathbf{x}$ in (1.1)
represents a linear continuous functional on $\mathbb{R}^{d}$. Conversely,
by Riesz representation theorem, every linear functional on $\mathbb{R}^{d}$
is of the form $\mathbf{w\cdot x}$, where $\mathbf{w}\in \mathbb{R}^{d}$ and
$\mathbf{x}=(x_{1},...,x_{d})$ is the variable (see \cite[Theorem 13.32]%
{Roman}). Linear continuous functionals constitute a significant subclass in
$C(\mathbb{R}^{d})$, denoted here by $\mathcal{L}(\mathbb{R}^{d})$. Thus,
UAT asserts that for certain activation functions $\sigma $ and any compact
set $K\subset \mathbb{R}^{d}$, the set%
\begin{equation*}
\mathcal{M}(\sigma )=span\{\sigma (f(x)-\theta ):f\in \mathcal{L}(\mathbb{R}%
^{d}),\theta \in \mathbb{R}\}
\end{equation*}%
is dense in $C(K)$. This observation tells the following generalization of
single hidden layer networks from $\mathbb{R}^{d}$ to any topological space $%
X,$ where $\mathcal{L}(\mathbb{R}^{d})$ is replaced with a fixed family of
functions (which need not be linear) from $C(X)$. We refer to such a family
as a \textit{basic family} for the feedforward neural networks with inputs
from a topological space (TFNNs). If $\mathcal{A}(X)\subset C(X)$ is a basic
family, then the architecture of a single hidden layer TFNN can be described
as follows:

\begin{itemize}
\item \textbf{Input Layer:} This layer consists of an element $x\in X$,
where $X$ is an arbitrary topological space.

\item \textbf{Hidden layer:} Each neuron in the hidden layer takes the input
$x$ from the input layer and applies a function $f\in \mathcal{A}(X)$ to $x$%
. This value is then multiplied by a weight $w$. A shift $\theta $ and then
a fixed activation function $\sigma :$ $\mathbb{R\rightarrow R}$ are applied
to $f(x)$. The resulting value $\sigma (wf(x)-\theta )$ represents the
output signal of the neuron.

\item \textbf{Output layer:} Each neuron in this layer receives weighted
signals from each neuron in the hidden layer, sums them up, and produces the
final output value.
\end{itemize}

This architecture represents a substantial generalization of classical
feedforward neural networks. Traditional networks are recovered only in the
specific case where $X=\mathbb{R}^{d}$ and $\mathcal{A}(X)=\mathcal{L}(%
\mathbb{R}^{d})$, yielding a structure with $d$ input neurons, each
processing a real-valued component $x_{1},x_{2},...,x_{d}\in \mathbb{R}$. In
the more general setting, however, the input $x\in X$ carries all
information processed by the input layer as a whole. This generalized
architecture allows the network to handle a wide variety of input types. In
general, a single hidden layer TFNN computes a function of the form%
\begin{equation*}
\sum_{i=1}^{r}c_{i}\sigma (w_{i}f_{i}(x)-\theta _{i}),\eqno(1.2)
\end{equation*}%
where $x\in X$ is the input, $f_{i}\in \mathcal{A}(X),$ $c_{i},w_{i},\theta
_{i}\in \mathbb{R}$ are the parameters of the network, and $\sigma :$ $%
\mathbb{R\rightarrow R}$ is a fixed activation function.

The aim of this paper is to show that for a broad a class of activation
functions $\sigma $, neural networks of the form presented in (1.2) can
approximate any continuous function on a compact subset $K\subset X$ with
arbitrary precision. In other words, the set
\begin{equation*}
\mathcal{N}(\sigma )=span\{\sigma (wf(x)-\theta ):f\in \mathcal{A}%
(X);~w,\theta \in \mathbb{R}\}
\end{equation*}%
is dense in $C(K)$ for every compact set $K\subset X$. As an application of
this result, we will derive an approximative version of the \textit{%
Kolmogorov superposition theorem} (KST) for compact metric spaces, where
outer (non-fixed and generally nonsmooth) functions are substituted with a
fixed ultimately smooth function.

\textbf{Related literature:} Extensive research has investigated the UAT
across various activation functions $\sigma $, examining how different
choices influence the approximation capabilities of neural networks. The
most general result in this area was obtained by Leshno, Lin, Pinkus and
Schocken \cite{Leshno}. They proved that a continuous activation function $%
\sigma $ possesses the UAP if and only if it is not a polynomial. This
result demonstrates the effectiveness of the single hidden layer perceptron
model across a wide range of activation functions $\sigma $. It should be
noted that, the universal approximation theorem in \cite{Leshno} was shown
to apply to a broader class of activation functions beyond just continuous
ones, including activation functions that may have discontinuities on sets
of Lebesgue measure zero. However, this paper will specifically concentrate
on continuous activation functions. For a thorough, step-by-step proof of
this theorem, refer to \cite{Pet,Pinkus}.

In the past, it was commonly accepted and highlighted in numerous studies
that attaining the universal approximation property necessitate large
networks with a substantial number of hidden neurons (see, e.g., \cite[%
Chapter 6.4.1]{Good}). In the above-mentioned earlier works, the number of
hidden neurons was regarded as unbounded. However, more recent research \cite%
{GI2,GI3,Ism} has demonstrated that neural networks using certain
non-explicit but practically computable activation functions can approximate
any continuous function over any compact set to any desired level of
accuracy, even with a minimal and fixed number of hidden neurons.

It should be noted that the UAP of neural networks operating between Banach
spaces has been explored in various studies. For example, in \cite{Sun}, the
fundamentality of ridge functions was established in a Banach space and
subsequently applied to shallow networks with a sigmoidal activation
function (see also \cite{Light}). Recall that a subset $S$ in a topological
vector space $E$ is said to be \textquotedblleft fundamental" if the linear
span of $S$ is dense in $E$. In \cite{Chen2}, the authors showed that any
continuous nonlinear function mapping a compact set $V$ in a Banach space of
continuous functions $C(K_{1})$ into $C(K_{2})$ can be approximated
arbitrarily well by shallow feedforward neural networks. Here $K_{1}$ and $%
K_{2}$ represent two compact sets in an abstract Banach space $X$ and the
Euclidean space $\mathbb{R}^{d}$, respectively. In \cite{Lu}, this approach
was extended to deep neural networks and referred to as DeepONet. In \cite%
{Lant}, the authors examined DeepONet within the context of an
encoder-decoder network framework, investigating its approximation
properties when the input space is a Hilbert space. In \cite{Kor},
quantitative estimates (i.e., convergence rates) for the approximation of
nonlinear operators using single-hidden layer networks in
infinite-dimensional Banach spaces were provided, extending some previous
results from the finite-dimensional case.

The UAP of infinite-dimensional neural networks, with inputs from Fr\'{e}%
chet spaces and outputs from Banach spaces, was established in \cite{Gal1}.
In \cite{Gal2}, the scope of this architecture was extended by proving
several universal approximation theorems for quasi-Polish input and output
spaces. The UAT for so-called functional input neural networks, defined on
weighted spaces with output values in a Banach space, was proved in \cite%
{Cuch}.

Modifications of traditional neural networks and neural network
architectures for mappings between non-Euclidean spaces, different from
those considered here, have been discussed in several other papers. For
example, in \cite{Krat1}, the authors addressed the question: \textit{Which
modifications to the input and output layers of a neural network preserve
its universal approximation capabilities?} They obtained sufficient
conditions on topological spaces $X$ and $Y$, along with a pair of feature
maps $\phi :X\rightarrow \mathbb{R}^{n}$ and $\rho :\mathbb{R}%
^{m}\rightarrow Y$, ensuring that if a family of functions $\mathcal{F}$ is
dense in $C(\mathbb{R}^{n},\mathbb{R}^{m})$, then the set $\left\{ \phi
\circ f\circ \rho :f\in \mathcal{F}\right\} $ is dense in $C(X,Y)$, where
density is understood with respect to the topology of uniform convergence on
compact sets. The set $\mathcal{F}$ represents an arbitrary expressive
neural network architecture. For example, $\mathcal{F}$ can be taken as the
set of single hidden layer networks with a continuous non-polynomial
activation function.

As a continuation of \cite{Krat1} and motivated by developments in the
mathematics of deep learning, the authors of \cite{Krat2} constructed
universal function approximators for continuous maps between arbitrary
Polish metric spaces $X$ and $Y$, using elementary functions between
Euclidean spaces as building blocks. In \cite{Acc}, a new geometric deep
learning model called geometric transformers was proposed. The authors
demonstrated that these models can approximate H\"{o}lder continuous functions $%
f:K\rightarrow Y$, where $K$ is a compact subset of $\mathbb{R}^{d}$ and $Y$
is a suitable metric space. In \cite{Krat3}, universal approximation
theorems were obtained for neural operators (NOs) and mixtures of neural
operators (MoNOs) acting between Sobolev spaces. More precisely, it was
shown that any non-linear continuous operator acting between Sobolev spaces $%
H^{s_{1}}$ and $H^{s_{2}}$ can be uniformly approximated over any compact
set $K\subset $ $H^{s_{1}}$ with arbitrary accuracy $\varepsilon $ using NOs
and MoNOs: $H^{s_{1}}\rightarrow H^{s_{2}}$. Moreover, the quantitative
results of \cite{Krat3} estimate the depth, width, and rank of the neural
operators in terms of the radius of $K$ and $\varepsilon $.

Recent research has demonstrated the universal approximation theorem (UAT)
for various hypercomplex-valued neural networks, including complex-,
quaternion-, tessarine-, and Clifford-valued networks, as well as more
general vector-valued neural networks (V-nets) defined over a
finite-dimensional algebra (see \cite{Valle} and references therein). We
hope that the results of this paper will stimulate further exploration of
these neural networks, particularly with outputs from these and other
general spaces.

\textbf{Comparison with other results:} To the best of our knowledge, this
is the first paper to explore the UAT for neural networks defined on a
general topological space. No additional structures, such as linearity,
metricity, completeness, or others, are required. The only restriction of
our main result is that the output space must be the usual real line. The
techniques utilized in this paper are not yet sufficient to generalize the
results to neural networks with outputs in infinite-dimensional spaces.
However, we hope this research will stimulate future work in this direction.

As indicated in the literature review above, all previous generalizations of
UAT involve general input-output correspondences where the input is a
topological space, always with additional structure, and the output is the
familiar $\mathbb{R}^{n}$, or the input space is the same as before, and the
output is also a topological space with additional structure. In most cases,
this additional structure consists of norms and inner products, which often
make the underlying spaces Banach or Hilbert spaces. Recent studies have
also involved Fr\'{e}chet spaces (complete and metrizable locally convex
topological vector spaces) and even quasi-Polish spaces (general topological
spaces that admit the existence of a countable family of real-valued
continuous functions that separate points).

Clearly, the problem of function approximation by neural networks with
outputs in $\mathbb{R}^{n}$ can be reduced to the case where the output lies
in $\mathbb{R}$. This is because any $\mathbb{R}^{n}$-valued function can be
treated as a vector of $n$ scalar-valued functions, each mapping into $%
\mathbb{R}$. Therefore, it suffices to construct a neural network that
approximates each scalar component individually, using a single neuron in
the output layer for each component.

Taking this into account, in the case of finite-dimensional outputs, our
results can be compared with earlier results that involve shallow networks
as follows. Our main result employs Tauber-Wiener activation functions, the
broadest class of activation functions for which UAT holds. These functions
automatically (by definition) guarantee approximation with arbitrary
accuracy in the univariate setting. Without this basic property, no UAT can
be proven in multivariate or more abstract settings. Some classical and
recent papers devoted to UAT in general spaces involve sigmoidal functions
(see \cite{Light, Sun}), weak sigmoidal functions --- those that tend to $0$
as $x\rightarrow $ $-\infty $, but not necessarily to $1$ as $x\rightarrow $
$+\infty $ (see \cite{Valle}) --- discriminatory functions (see \cite{Gal1}%
), or activation functions with a separating property (see \cite{Gal2}),
which is the infinite-dimensional counterpart to the well-known sigmoidal
property for real univariate functions. The work \cite{Chen2} considers
Tauber-Wiener functions, but the underlying input space is Banach, which is
far from a general topological space. Note that the mildest known condition
imposed on a topological space is \textquotedblleft quasi-Polish" (see \cite%
{Gal2}).

In summary, we claim that if the output space is $\mathbb{R}^{n}$, then the
UAT proven in this paper is more general with respect to both the input
space and/or activation functions compared to classical and the recent
corresponding results.

We also apply our main result to obtain an approximate version of
Kolmogorov's superposition theorem for compact metric spaces. Note that
although much research has focused on the role of the classical Kolmogorov
theorem in conventional neural networks, and many studies have been devoted
to the generalization of neural networks to infinite-dimensional spaces, the
extension of Kolmogorov's theorem (due to Ostrand) has not yet been
considered. In Ostrand's theorem, the outer functions depend on the target
function and may be poorly behaved. Our second result shows that these outer
functions can be replaced with a single, ultimately smooth, universal
function (not depending on the target function) to approximate every
continuous function on a compact metric space with arbitrary precision.

\bigskip

\begin{center}
{\large \textbf{2. Main results}}
\end{center}

In this section, we analyze the conditions under which shallow networks with
inputs from a topological space possess the universal approximation property.

Assume $X$ is an arbitrary topological space. In the sequel, in $C(X)$, we
will use the topology of uniform convergence on compact sets. This topology
is induced by the seminorms
\begin{equation*}
\left\Vert g\right\Vert _{K}=\max_{x\in K}\left\vert g(x)\right\vert ,
\end{equation*}%
where $K$ are compact sets in $X$. A subbasis at the origin for this
topology is given by the sets%
\begin{equation*}
U(K,r)=\left\{ g\in C(X):\left\Vert g\right\Vert _{K}<r\right\} ,
\end{equation*}%
where $K\subset X$ is compact and $r>0$. A sequence (or net) $\{g_{n}\}$ in
this topology converges to $g$ iff $\left\Vert g_{n}-g\right\Vert
_{K}\rightarrow 0$ for every compact set $K\subset X$. Thus, in what
follows, when we say that $B$ is dense in $C(X)$, we will mean that $B$ is
dense with respect to the aforementioned topology of uniform convergence on
compact sets.

We say that a subclass $\mathcal{A}(X)\subset C(X)$ holds the $D$-property
if the set

\begin{equation*}
S=span\left\{ u\circ v:u\in C(\mathbb{R}),v\in \mathcal{A}(X)\right\} \eqno%
(2.1)
\end{equation*}%
is dense in $C(X)$.

In what follows, we will use activation functions $\sigma :\mathbb{R}%
\rightarrow \mathbb{R}$ (whether continuous or discontinuous) with the
property that the $span\{\sigma (wx-\theta ):w\in \mathbb{R},\theta \in
\mathbb{R}\}$ is dense in every $C[a,b]$. Such functions are called
Tauber-Wiener (TW) functions (see \cite{Chen2}).

\bigskip

\textbf{Theorem 2.1.} \textit{Assume $X$ is a topological space, $\mathcal{A}%
(X)$ is a subclass of $C(X)$ with the $D$-property and $\sigma :\mathbb{R}%
\rightarrow \mathbb{R}$ is a TW function. Then for any $\varepsilon >0$, any
compact set $K\subset X$ and any function $g\in C(X)$ there exist $r\in
\mathbb{N}$, $f_{i}\in \mathcal{A}(X)$, $c_{i},w_{i},\theta _{i},\in \mathbb{%
R}$, $i=1,...,r$, such that}
\begin{equation*}
\max_{x\in K}\left\vert g(x)-\sum_{i=1}^{r}c_{i}\sigma (w_{i}f_{i}(x)-\theta
_{i})\right\vert <\varepsilon .
\end{equation*}%
\textit{That is, TFNNs with inputs from $X$ is dense in $C(X)$.}

\bigskip

\textbf{Proof.} Take any $\varepsilon >0$, any compact set $K\subset X$ and
any function $g\in C(X)$. Since $\mathcal{A}(X)$ has the $D$-property, there
exist finitely many functions $u_{i}\in C(\mathbb{R})$ and $v_{i}\in \mathcal{A}(X)$
such that
\begin{equation*}
\left\vert g(x)-\sum_{i=1}^{n}u_{i}(v_{i}(x))\right\vert <\varepsilon /2,%
\eqno(2.2)
\end{equation*}%
for all $x\in K$.

Since $v_{i}$ are continuous, the images $v_{i}(K)$ are compact sets in $%
\mathbb{R}$. Set $V=\cup _{i=1}^{n}v_{i}(K)$. Note that $V$ is also compact.

Since $\sigma $ is a TW function, each continuous univariate function $%
u_{i}(t)$, $t$ $\in V$, can be approximated by single hidden layer networks
with the activation function $\sigma $. Thus, there exist coefficients $%
c_{ij},w_{ij},\theta _{ij}\in \mathbb{R}$, $1\leq i\leq n$, $1\leq j\leq
k_{i}$, such that

\begin{equation*}
\left\vert u_{i}(t)-\sum_{j=1}^{k_{i}}c_{ij}\sigma (w_{ij}t-\theta
_{ij})\right\vert <\varepsilon /2n
\end{equation*}%
for all $t\in V$. Therefore,

\begin{equation*}
\left\vert u_{i}(v_{i}(x))-\sum_{j=1}^{k_{i}}c_{ij}\sigma
(w_{ij}v_{i}(x)-\theta _{ij})\right\vert <\varepsilon /2n\eqno(2.3)
\end{equation*}%
for each $i=1,...,n$, and all $x\in K$. It follows from (2.2) and (2.3) that

\begin{equation*}
\left\vert g(x)-\sum_{i=1}^{n}\sum_{j=1}^{k_{i}}c_{ij}\sigma
(w_{ij}v_{i}(x)-\theta _{ij})\right\vert <\varepsilon
\end{equation*}%
for any $x\in K$. This completes the proof of Theorem 2.1.

\bigskip

\textbf{Remark}. Theorem 2.1 generalizes existing universal approximation
theorems for traditional feedforward neural networks. In such networks, the
space of linear continuous functionals on $\mathbb{R}^{d}$ serves as the
basic family $\mathcal{A}(X)$, which clearly satisfies the $D$-property.
Indeed, linear continuous functionals on $\mathbb{R}^{d}$ are of the form $%
\mathbf{a}\cdot \mathbf{x}$, where $\mathbf{a}\in \mathbb{R}^{d}$, $\mathbf{x%
}\in \mathbb{R}^{d}$, and $\cdot $ denotes the scalar product. Since the
linear span of ridge functions $g(\mathbf{a}\cdot \mathbf{x)}$ is dense in $%
C(\mathbb{R}^{d})$ in the topology of uniform convergence on compact sets
(see, e.g., \cite[p. 12]{Ism}), $D$-property is satisfied in this setting.

\bigskip

Note that, in particular, $X$ may be a topological vector space. For such a
space, $X^{\ast }$ denotes the continuous dual of $X$, which is the space of
linear continuous functionals defined on $X$. The following theorem is based
on Theorem 2.1.

\bigskip

\textbf{Theorem 2.2.} \textit{Assume $X$ is a locally convex topological
vector space (in particular, a normed space) and $\sigma $ is a continuous
univariate function that is not a polynomial. Then for any $\varepsilon >0$,
any compact set $K\subset X$ and any function $g\in C(K)$ there exist $r\in
\mathbb{N}$, $f_{i}\in X^{\ast }$, $c_{i},\theta _{i}\in \mathbb{R}$, $%
i=1,...,r$, such that}
\begin{equation*}
\max_{x\in K}\left\vert g(x)-\sum_{i=1}^{r}c_{i}\sigma (f_{i}(x)-\theta
_{i})\right\vert <\varepsilon .
\end{equation*}

\bigskip

The proof of this theorem relies on Theorem 2.1 and the following two facts.

\textit{Fact 1.} The space $X^{\ast }$ possesses the $D$-property.

Let us prove this fact. Specifically, this property holds if in (2.1),
instead of all $u\in C(\mathbb{R})$, we take the single function $u(t)=e^{t}$%
. That is, we claim that the set

\begin{equation*}
S=span\{e^{r(x)}:r\in X^{\ast }\}.
\end{equation*}%
is dense in $C(K)$ for every compact set $K\subset X$.

Indeed, first it is not difficult to see that $S$ is a subalgebra of $C(X)$.
To see this, note that for any $r_{1},r_{2}\in X^{\ast }$
\begin{equation*}
e^{r_{1}(x)}e^{r_{2}(x)}=e^{r_{1}(x)+r_{2}(x)}\in E\text{,}
\end{equation*}%
since $r_{1}+r_{2}\in X^{\ast }$. Therefore, the linear space $S$ is closed
under multiplication, indicating that $S$ is an algebra.

Second, if $r$ is the zero functional, then $e^{r(x)}=1$, showing that $S$
contains all constant functions.

Now since $X$ is locally convex, the Hahn-Banach extension theorem holds. It
is a consequence of this theorem that for any distinct points $%
x_{1},x_{2}\in X$ there exists a functional $r\in X^{\ast }$ such that $%
r(x_{1})\neq r(x_{2})$ (see \cite[Theorem 3.6]{Rudin}). Hence, the algebra $%
S $ separates points in $X$. By the Stone-Weierstrass theorem \cite{Stone},
for any compact $K\subset X$, the algebra $S$ restricted to $K$ is dense in $%
C(K)$. In other words, the space $X^{\ast }$ has the $D$-property.

\textit{Fact 2.} A continuous nonpolynomial activation function $\sigma $ is
a TW function.

This fact follows from the main result of \cite{Leshno} that a continuous
nonpolynomial activation function provides the universal approximation
property for traditional single hidden layer networks.

\bigskip

Let us now we apply Theorem 2.1 to derive an approximative version of the
renowned Kolmogorov superposition theorem (KST) for compact metric spaces.
KST \cite{Kol} states that for the unit cube $\mathbb{I}^{d},~\mathbb{I}%
=[0,1],~d\geq 2,$ there exist $2d+1$ functions $\{s_{q}\}_{q=1}^{2d+1}%
\subset C(\mathbb{I}^{d})$ of the form
\begin{equation*}
s_{q}(x_{1},...,x_{d})=\sum_{p=1}^{d}\varphi _{pq}(x_{p}),~\varphi _{pq}\in
C(\mathbb{I}),~p=1,...,d,~q=1,...,2d+1,
\end{equation*}%
such that each function $f\in C(\mathbb{I}^{d})$ admits the representation
\begin{equation*}
f(\mathbf{x})=\sum_{q=1}^{2d+1}g_{q}(s_{q}(\mathbf{x})),~\mathbf{x}%
=(x_{1},...,x_{d})\in \mathbb{I}^{d},~g_{q}\in C({{\mathbb{R)}}}.
\end{equation*}

This surprising and deep result, which solved (negatively) Hilbert's 13-th
problem, has been improved and generalized in several directions. For
detailed information about KST, including its refinements, various variants,
and generalizations, see the monographs \cite[Chapter 1]{Kh} and \cite[%
Chapter 4]{Ism}. It is worth noting that KST initiated a new line of
research in approximation theory --- namely, the approximation of
multivariate functions by various combinations of functions of fewer
variables (see, e.g., \cite{Bab1,Bab2,Gar,Gol,Is,Tem}). The relevance of KST
to neural networks, along with its theoretical and computational aspects,
has been extensively discussed in the neural network literature (see, e.g.,
\cite{AV} and references therein).

Ostrand \cite{Ost} extended KST to general compact metric spaces as follows.

\bigskip

\textbf{Theorem 2.3.} (Ostrand \cite{Ost}). \textit{For $p=1,2,...,n$, let $%
X_{p}$ be a compact metric space of finite dimension $d_{p}$ and let $%
m=\sum_{p=1}^{n}d_{p}.$ There exist universal continuous functions $\psi
_{pq}:X_{p}\rightarrow \lbrack 0,1],$ $p=1,...,n,$ $q=1,...,2m+1,$ such that
every continuous function $g$ defined on $\Pi _{p=1}^{n}X_{p}$ is
representable in the form}
\begin{equation*}
g(x_{1},...,x_{n})=\sum_{q=1}^{2m+1}h_{q}\left( \sum_{p=1}^{n}\psi
_{pq}(x_{p})\right) ,
\end{equation*}%
\textit{where $h_{q}$ are continuous functions depending on $g$.}

\bigskip

Although substantial research has been conducted on generalizing neural
networks to abstract spaces, Ostrand's theorem has not yet been considered
in this context. Like Kolmogorov's original superposition theorem, Ostrand's
result can also be interpreted as a special case of a two-hidden-layer
feedforward neural network model, where the inputs lie in a general compact
metric space. We refer the reader to \cite{AV} for such a model when the
input space is $\mathbb{R}^{n}$. From an application perspective, however,
Theorem 2.3 is not suitable, as the outer functions $h_{q}$ are not fixed
and depend on the target function $g$. In standard feedforward neural
network theory, activation functions are typically fixed and do not depend
on the functions being approximated. Our next theorem demonstrates that
Ostrand's result can be adapted to the neural network framework by replacing
the outer functions $g_{q}$, which may be highly non-smoth, with a single
infinitely differentiable universal activation function $\sigma $. This
yields a model capable of approximating any continuous function defined on a
compact metric space. The proof relies on Theorem 2.1 and the construction
of specific activation functions, which we refer to as superactivation
functions.

A function $\sigma \in C(\mathbb{R})$ with the property that, for any
interval $[a,b]\subset \mathbb{R}$, the set $\Lambda =\{\sigma (wx-\theta
):w,\theta \in \mathbb{R}\}$ is dense in $C[a,b]$ is called a \textit{%
superactivation function}. Note that here $\Lambda $ is not the linear span
of the functions $\sigma (wx-\theta )$, but rather a very narrow subclass of
it. Superactivation functions indeed exist, which shows the following
proposition.

\bigskip

\textbf{Proposition 2.1.} \textit{There exist an infinitely differentiable
superactivation function $\sigma $. Moreover, such a function can be
constructed algorithmically.}

\bigskip

\textbf{Proof.} Let $\alpha $ be any positive real number. Divide the
interval $[\alpha ,+\infty )$ into the segments $[\alpha ,2\alpha ],$ $%
[2\alpha ,3\alpha ],...$. Let $\{p_{n}(t)\}_{n=1}^{\infty }$ be the sequence
of polynomials with rational coefficients defined on $[0,1].$ We construct $%
\sigma $ in two stages. In the first stage, we define $\sigma $ on the
closed intervals $[(2m-1)\alpha ,2m\alpha ],$ $m=1,2,...$ as the function
\begin{equation*}
\sigma (t)=p_{m}(\frac{t}{\alpha }-2m+1),\text{ }t\in \lbrack (2m-1)\alpha
,2m\alpha ],
\end{equation*}%
or equivalently,
\begin{equation*}
\sigma (\alpha t+(2m-1)\alpha )=p_{m}(t),\text{ }t\in \lbrack 0,1].\eqno(2.4)
\end{equation*}%
In the second stage, we extend $\sigma $ to the intervals $(2m\alpha
,(2m+1)\alpha ),$ $m=1,2,...,$ and $(-\infty ,\alpha )$, maintaining the $%
C^{\infty }$ property. To do this we use the smooth transition function

\begin{equation*}
\beta (t)=\left\{
\begin{array}{c}
e^{-\frac{1}{t}}\text{ if }t>0, \\
0\text{ if }t\leq 0,%
\end{array}%
\right.
\end{equation*}%
Then the function

\begin{equation*}
\omega (t):=\frac{\beta (\frac{t-2m\alpha }{\alpha })}{\beta (\frac{%
t-2m\alpha }{\alpha })+\beta (1-\frac{t-2m\alpha }{\alpha })}
\end{equation*}%
has the following properties

1) $\omega (t)=0$ for $t\leq 2m\alpha ,$

2) $\omega (t)=1$ for $t\geq (2m+1)\alpha ,$

3) $\omega (t)\in (0,1)$ for $t\in (2m\alpha ,(2m+1)\alpha ),$

4) $\omega \in C^{\infty }(\mathbb{R}).$

Now the function

\begin{equation*}
\tau(t):=(1-\omega(t))p_m\left(\frac{t}{\alpha}-2m+1\right)+\omega(t)p_{m+1}\left(\frac{t}{\alpha}-2m-1\right)
\end{equation*}
provides a smooth extension of $\sigma$ to the interval $(2m\alpha,(2m+1)\alpha)$. By the similar way, we can
smoothly extend $\sigma $ to the half line $(-\infty ,\alpha )$.

For any univariate function $h\in C[0,1]$ and any $\varepsilon >0$ there
exists a polynomial $p(t)$ with rational coefficients such that
\begin{equation*}
\left\vert h(t)-p(t)\right\vert <\varepsilon ,
\end{equation*}%
for all $t\in \lbrack 0,1].$ This together with (2.4) mean that
\begin{equation*}
\left\vert h(t)-\sigma (\alpha t-s)\right\vert <\varepsilon ,\eqno(2.5)
\end{equation*}%
for some $s\in \mathbb{R}$ and all $t\in \lbrack 0,1].$

Using linear transformation it is not difficult to go from $[0,1]$ to any
finite closed interval $[a,b]$. Indeed, let $u\in C[a,b]$, $\sigma $ be
constructed as above and $\varepsilon $ be an arbitrarily small positive
number. The transformed function $h(t)=u(a+(b-a)t)$ is well defined on $%
[0,1] $ and we can apply inequality (2.5). Now using the inverse
transformation $t=\frac{x-a}{b-a}$, we can write that
\begin{equation*}
\left\vert u(x)-\sigma (wx-\theta )\right\vert <\varepsilon ,\eqno(2.6)
\end{equation*}%
for all $x\in \lbrack a,b]$, where $w=\frac{\alpha }{b-a}$ and $\theta =%
\frac{\alpha a}{b-a}+s$.

Inequality (2.6) proves the proposition.

\bigskip

Superactivation functions demonstrate that shallow networks can approximate
univariate continuous functions with the minimal number of hidden neurons;
in fact, a single hidden neuron is sufficient. Similar activation functions $%
\sigma $, with additional properties of monotonicity and sigmoidality, were
algorithmically constructed in \cite{GI2} and utilized in practical
examples. It should be remarked that the existence of activation functions
that ensure universal approximation for single and two hidden layer neural
networks with a fixed number of hidden units was first established in \cite%
{Mai}.

\bigskip

\textbf{Theorem 2.4.} \textit{For $p=1,2,...,n$, let $X_{p}$ be a compact
metric space of finite dimension $d_{p}$ and let $m=\sum_{p=1}^{n}d_{p}.$
Assume $\sigma :\mathbb{R}\rightarrow \mathbb{R}$ is a superactivation
function. Then there exist universal continuous functions $\psi
_{pq}:X_{p}\rightarrow \lbrack 0,1],$ $p=1,...,n,$ $q=1,...,2m+1,$ such that
for every continuous function $g$ defined on $X=\Pi _{p=1}^{n}X_{p}$ and any
$\varepsilon >0$ there exist $w_{q},\theta _{q},\in \mathbb{R}$, $%
q=1,...,2m+1$, satisfying the inequality}%
\begin{equation*}
\left\vert g(x_{1},...,x_{n})-\sum_{q=1}^{2m+1}\sigma \left(
w_{q}\sum_{p=1}^{n}\psi _{pq}(x_{p})-\theta _{q}\right) \right\vert
<\varepsilon ,
\end{equation*}%
\textit{for all $(x_{1},...,x_{n})\in X$.}

\bigskip

The proof is based on the preceding material. It follows from Theorem 2.3
that for the metric space $X=\Pi _{p=1}^{n}X_{p}$, the family of $2m+1$
functions
\begin{equation*}
\mathcal{K}(X)=\left\{ \sum_{p=1}^{n}\psi _{pq}(x_{p}):q=1,...,2m+1\right\}
\end{equation*}%
satisfies the $D$-property in $C(X)$. Now, if in Theorem 2.1 we take $%
\mathcal{A}(X)=\mathcal{K}(X)$ and choose any superactivation function $%
\sigma $, then the number of terms $r$ will be $2m+1$. To see this, it is
sufficient to repeat the proof of Theorem 2.1, noting that $n=2m+1$, $%
k_{i}=1 $ and all $c_{ij}=1$. This observation completes the proof of
Theorem 2.4.

Note that in Theorem 2.4 the outer function $\sigma $ does not depend on $g$%
. Moreover, by Proposition 2.1, $\sigma $ can be chosen to be infinitely
differentiable. The only parameters that depend on $g$ are the numbers $%
\theta _{q}$. The numbers $w_{q}$ can be taken to be equal and fixed once
and for all. This is evident from the construction of $\sigma $ above (see
(2.6), where $w$ is fixed for all $u$). For example, if we set $\alpha =b-a$%
, where $[a,b]$ is a closed interval containing all the sets $\Psi _{q}(X),$
where $\Psi _{q}(x_{1},...,x_{n})=\sum_{p=1}^{n}\psi _{pq}(x_{p})$, $%
q=1,...,2m+1$, then $w_{q}$ can all be taken to be equal to $1$.

\bigskip

\textbf{Acknowledgements.} The author would like to thank the anonymous
reviewers for their careful reading of the manuscript and their insightful
comments and suggestions, which have helped to improve the clarity and
quality of the paper.

\smallskip

\textbf{Author Contributions.} Vugar Ismailov is the sole author who wrote the whole manuscript.

\smallskip

\textbf{Funding.} No funding was received for this study.

\smallskip

\textbf{Data Availability.} No datasets were generated or analysed during the current study.

\smallskip

\textbf{Declaration of interest statement:} The author declares that he has
no competing financial interests or personal relationships that could have
appeared to influence the work reported in this paper.

\end{document}